\documentclass[a4paper,12pt]{article}

\usepackage{float}
\usepackage{graphicx}
\usepackage[left=1.5in,right=1.5in,top=1in,bottom=1in]{geometry}
\usepackage[numbers, compress]{natbib}
\usepackage{amsmath}
\usepackage{mathrsfs}
\usepackage{amsfonts}
\usepackage{enumitem}
\usepackage{hyperref}
\usepackage{xcolor}

\fontfamily{qbk}\selectfont
\setlist[itemize]{leftmargin=*}
\newcommand\YUGE{\fontsize{30}{36}\selectfont}

\begin{document}

\raggedright

\begin{center}
\makebox[\textwidth][c]{\YUGE \textbf{A New Angle on L2 Regularization}}
%\\
%\vspace{-0.5cm}
\makebox[\textwidth][c]{\footnotesize (\textit{interactive version available at} \url{https://thomas-tanay.github.io/post--L2-regularization/})}
\end{center}

\vspace{-0.2cm}

\begin{figure}[h!]
  \centering
	\makebox[\textwidth][c]{\includegraphics[width=1.3\textwidth]{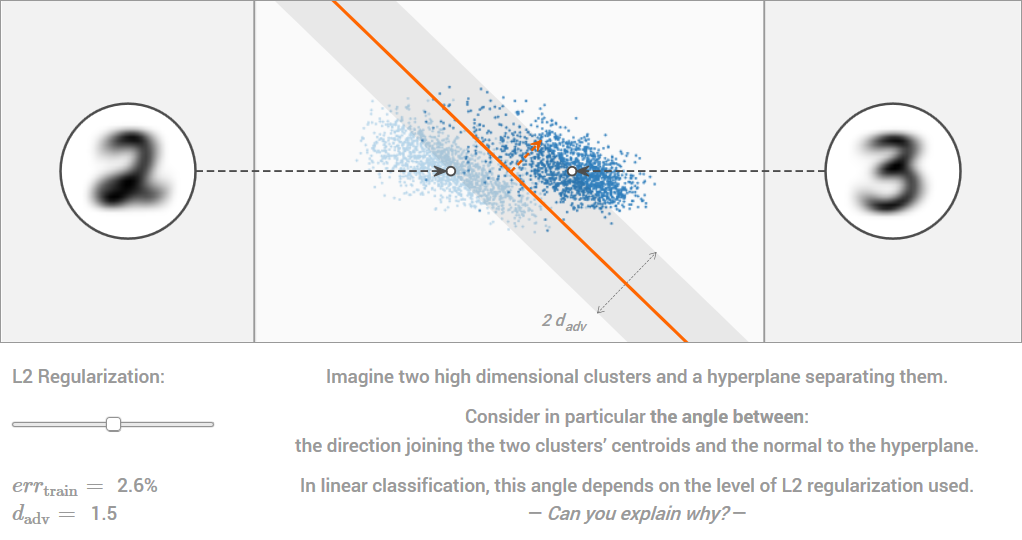}}
\end{figure}

\vspace{-0.6cm}

\noindent\makebox[\linewidth]{\textcolor{gray}{\rule{1.3\textwidth}{0.025cm}}}	

\vspace{-0.6cm}
{
\begin{center}
\begin{tabular}{ l l }
 {\footnotesize \textbf{Thomas Tanay}} & \quad{\footnotesize \textbf{Lewis D Griffin}} \\ 
 {\footnotesize \textbf{CoMPLEX, UCL}} & \quad{\footnotesize \textbf{CoMPLEX, UCL}}  
\end{tabular}
\end{center}}

Deep neural networks have been shown to be vulnerable to the adversarial example phenomenon: all models tested so far can have their classifications dramatically altered by small image perturbations \citep{szegedy2013intriguing,goodfellow2014explaining}. The following predictions were for instance made by a state-of-the-art network trained to recognize celebrities \citep{parkhi2015deep}:

\begin{figure}[h!]
  \centering
	\makebox[\textwidth][c]{\includegraphics[width=\textwidth]{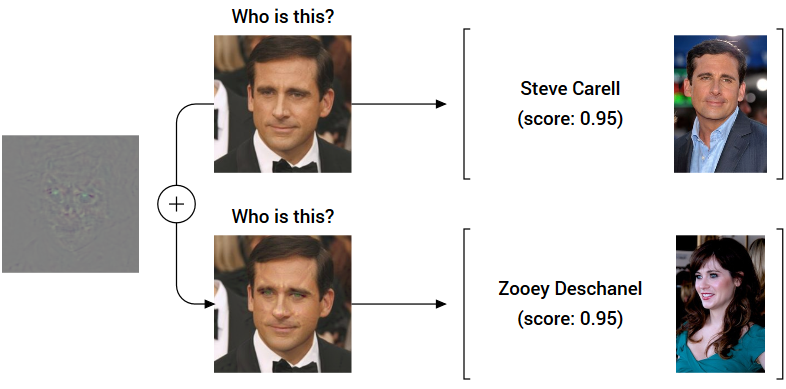}}
\end{figure}

This result is puzzling for two reasons. First, it challenges a common belief according to which good generalization to novel data and robustness to small perturbations go hand in hand. Second, it constitutes a potential threat to real-world applications \citep{papernot2016practical,kurakin2016adversariala,evtimov2017robust}. Researchers at MIT have for instance recently constructed 3D objects that are misclassified under a wide distribution of angles and viewpoints \citep{athalye2017synthesizing}. Understanding this phenomenon and improving deep networks' robustness has thus become an important research objective.

Several approaches have been explored already. The phenomenon has been described in detail \citep{moosavi2016deepfool,carlini2016towards} and some theoretical analysis has been provided \citep{bastani2016measuring,fawzi2016robustness,carlini2017ground}. Attempts have been made at designing more robust architectures \citep{gu2014towards,papernot2016distillation,zhao2016suppressing,rozsa2016towards} or at detecting adversarial examples during evaluation \citep{bhagoji2017dimensionality,feinman2017detecting,grosse2017statistical,metzen2017detecting}. \emph{Adversarial training} has also been introduced as a new regularization technique penalising adversarial directions \citep{goodfellow2014explaining,kurakin2016adversarialb,tramer2017ensemble,madry2017towards}. Unfortunately, the problem remains largely unresolved \citep{goodfellow2017attacking,carlini2017adversarial}. Confronted with this difficulty, we propose to proceed from fundamentals: focusing on linear classification first and then increasing complexity incrementally.
	
\noindent\makebox[\linewidth]{\textcolor{gray}{\rule{\paperwidth}{0.025cm}}}

\vspace{0.75cm}

{\Large A Toy Problem}

In linear classification, adversarial perturbations are often understood as a property of the dot product in high dimension. A widespread intuition is that: ``for high dimensional problems, we can make many infinitesimal changes to the input that add up to one large change to the output'' \citep{goodfellow2014explaining}. Here, we challenge this intuition and argue instead that adversarial examples exist when the classification boundary lies close to the data manifold---independently of the image space dimension.\footnote{This idea was originally inspired by the work of Marron et al. \citep{marron2007distance} on the \emph{data piling phenomenon} affecting SVMs on high-dimensional low sample size data.}

{\textbf{Setup}}

Let's start with a minimal toy problem: a two-dimensional image space where each image is a function of $a$ and $b$.

\vspace{0.5cm}

\begin{figure}[h!]
\includegraphics[width=0.95\textwidth]{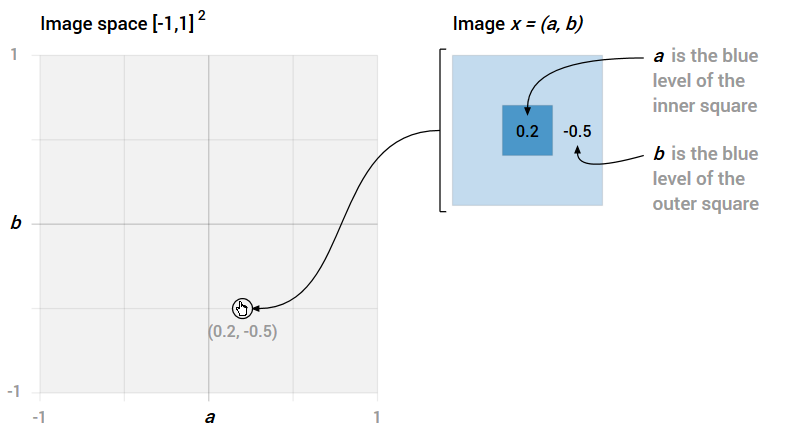}
\end{figure}

In this simple image space, we define two classes of images...

\begin{figure}[h!]
\includegraphics[width=0.95\textwidth]{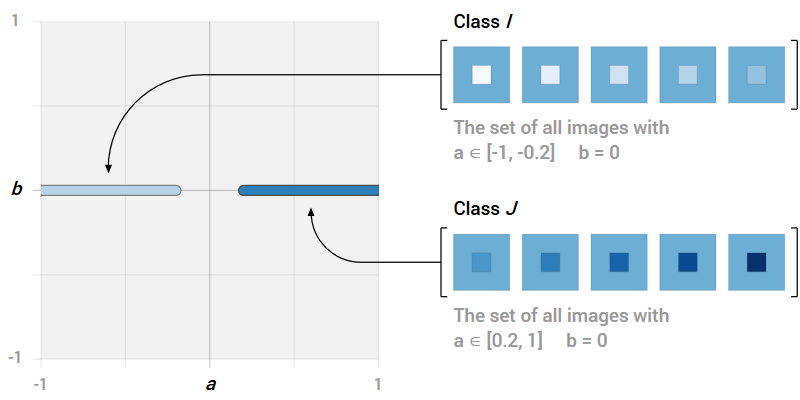}
\end{figure}

...which can be separated by an infinite number of linear classifiers. Consider for instance the line $\mathscr{L}_{\theta}$.

\begin{figure}[h!]
\includegraphics[width=0.95\textwidth]{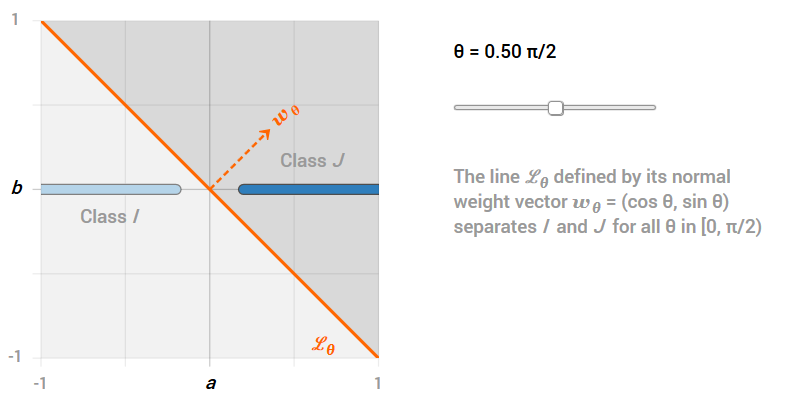}
\end{figure}

This raises a first question: if all the linear classifiers $\mathscr{L}_{\theta}$ separate $I$ and $J$ equally well, are they all equally robust to image perturbations?

{\textbf{Projected and mirror images}}

Consider an image $\boldsymbol{x}$ in class $I$. The closest image classified in the opposite class is \emph{the projected image} of $\boldsymbol{x}$ on $\mathscr{L}_{\theta}$:

\begin{figure}[h!]
\includegraphics[width=0.7\textwidth]{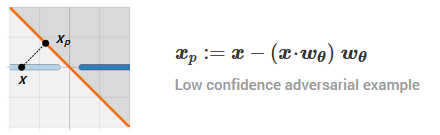}
\end{figure}

When $\boldsymbol{x}$ and $\boldsymbol{x}_p$ are very close to each other, we say that $\boldsymbol{x}_p$ is an \emph{adversarial example} of $\boldsymbol{x}$. Observe though that $\boldsymbol{x}_p$ is classified with a low confidence score (it lies on the boundary) and it is perhaps more interesting to consider \emph{high-confidence adversarial examples} \citep{carlini2017adversarial}.

In the following, we focus on the \emph{mirror image} of $\boldsymbol{x}$ through $\mathscr{L}_{\theta}$:

\begin{figure}[h!]
\includegraphics[width=0.7\textwidth]{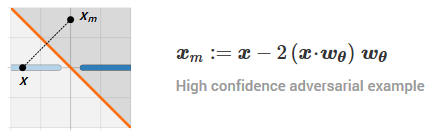}
\end{figure}

By construction, $\boldsymbol{x}$ and $\boldsymbol{x}_m$ are at the same distance from the boundary and are classified with the same confidence level.

{\textbf{A mirror image as a function of $\theta$}}

Coming back to our toy problem, we can now plot an image $\boldsymbol{x}$ and its mirror image $\boldsymbol{x}_m$ as a function of $\theta$.

\begin{figure}[h!]
\includegraphics[width=0.95\textwidth]{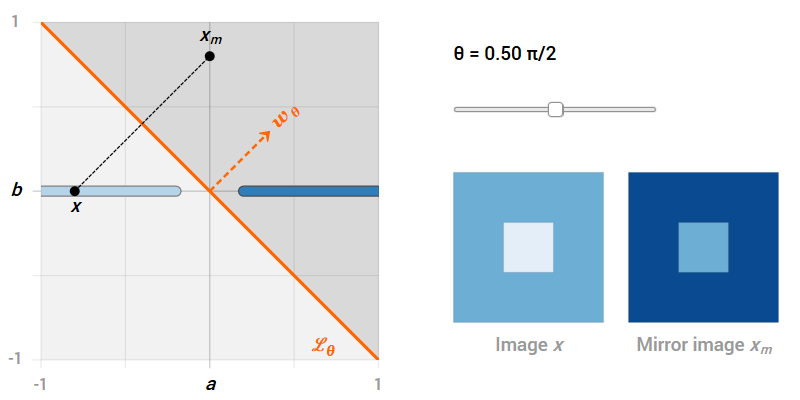}
\end{figure}

We see that the distance between $\boldsymbol{x}$ and $\boldsymbol{x}_m$ depends on the angle $\theta$. The two borderline cases are of particular interest.

\begin{figure}[h!]
\includegraphics[width=\textwidth]{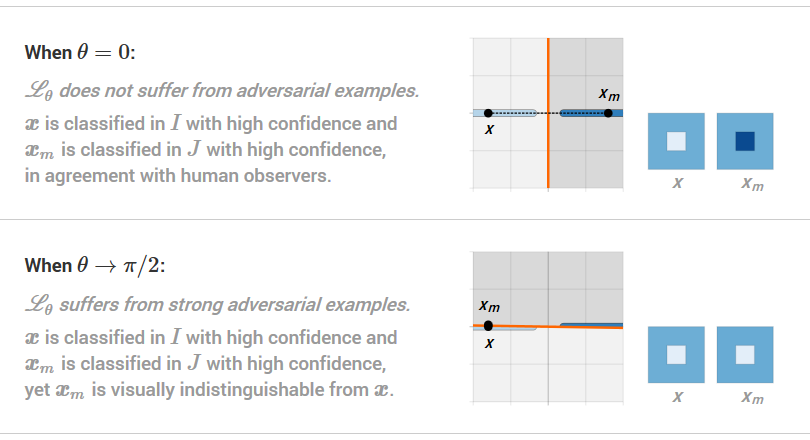}
\end{figure}

This raises a second question: if adversarial examples exist when $\mathscr{L}_{\theta}$ is strongly tilted, what makes $\mathscr{L}_{\theta}$ tilt in practice?

{\textbf{Overfitting and L2 regularization}}

Our working hypothesis is that the classification boundary defined by standard linear learning algorithms tilts by overfitting noisy data points in the training set. This hypothesis is supported by the theoretical result of Xu et al. \citep{xu2009robustness} relating robustness to regularization in Support Vector Machines (SVM). It can also be tested experimentally: techniques designed to reduce overfitting such as L2 regularization are expected to mitigate the adversarial example phenomenon.

Consider for instance a training set containing one noisy data point $\boldsymbol{p}$.

\begin{figure}[h!]
\includegraphics[width=0.95\textwidth]{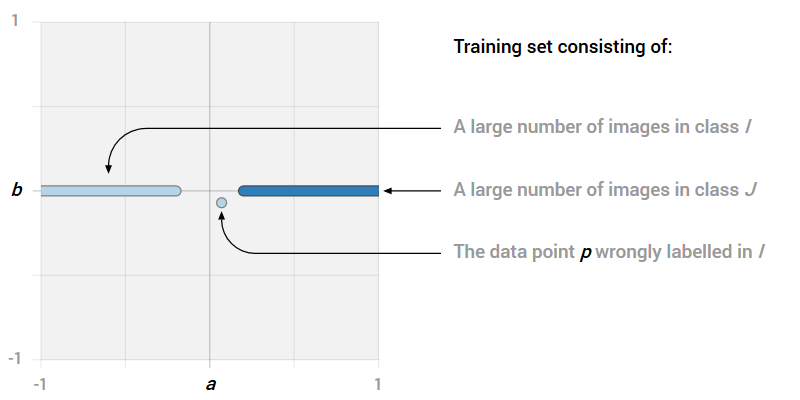}
\end{figure}

If we train an SVM or a logistic regression model on this training set, we observe two possible behaviours.

\begin{figure}[h!]
\includegraphics[width=\textwidth]{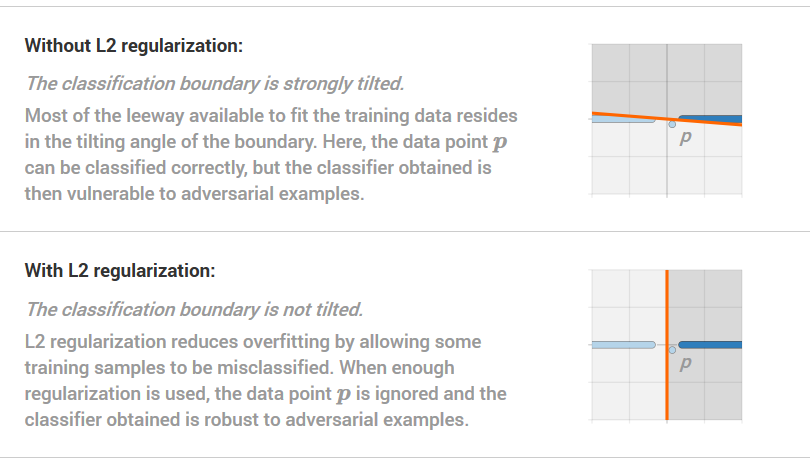}
\end{figure}

At this point, one might legitimately wonder---what does a 1-dimensional data manifold lying in a 2-dimensional image space have to do with high-dimensional natural images?

\noindent\makebox[\linewidth]{\textcolor{gray}{\rule{\paperwidth}{0.025cm}}}

\vspace{0.75cm}

{\Large Adversarial Examples in Linear Classification}

In the following, we show that the two main ideas introduced in the previous toy problem stay valid in the general case: adversarial examples exist when the classification boundary lies close to the data manifold and L2 regularization controls the tilting angle of the boundary.

\vspace{0.5cm}

{\Large \textit{Scaling the Loss Function}}

Let's start with a simple observation: during training, the norm of the weight vector acts as a scaling parameter on the loss function.

{\textbf{Setup}}

Let $I$ and $J$ be two classes of images and $\mathcal{C}$ a hyperplane boundary defining a linear classifier in $\mathbb{R}^d$. $\mathcal{C}$ is specified by a normal weight vector $\boldsymbol{w}$ and a bias $b$. For an image $\boldsymbol{x}$ in $\mathbb{R}^d$, we call \emph{raw score} of $\boldsymbol{x}$ through $\mathcal{C}$ the value:
$$s(\boldsymbol{x}) := \boldsymbol{w} \!\cdot\! \boldsymbol{x} + b$$
The raw score can be seen as a \emph{signed distance} between $\boldsymbol{x}$ and the classification boundary defined by $\mathcal{C}$. In particular:
$$\boldsymbol{x} \text{ is classified in } \mathrel{\Bigg|} \begin{array}{@{}c@{}} \text{$I$ if } s(\boldsymbol{x}) \leq 0 \\ \text{$J$ if } s(\boldsymbol{x}) \geq 0 \end{array}$$

Now, consider a training set $T$ of $n$ pairs $(\boldsymbol{x},y)$ where $\boldsymbol{x}$ is an image and $y = \{ -1 \text{ if } \boldsymbol{x} \in I \;|\; 1 \text{ if } \boldsymbol{x} \in J\}$ is its label. We are interested in the distributions of the following quantities over $T$:

\begin{figure}[h!]
\includegraphics[width=\textwidth]{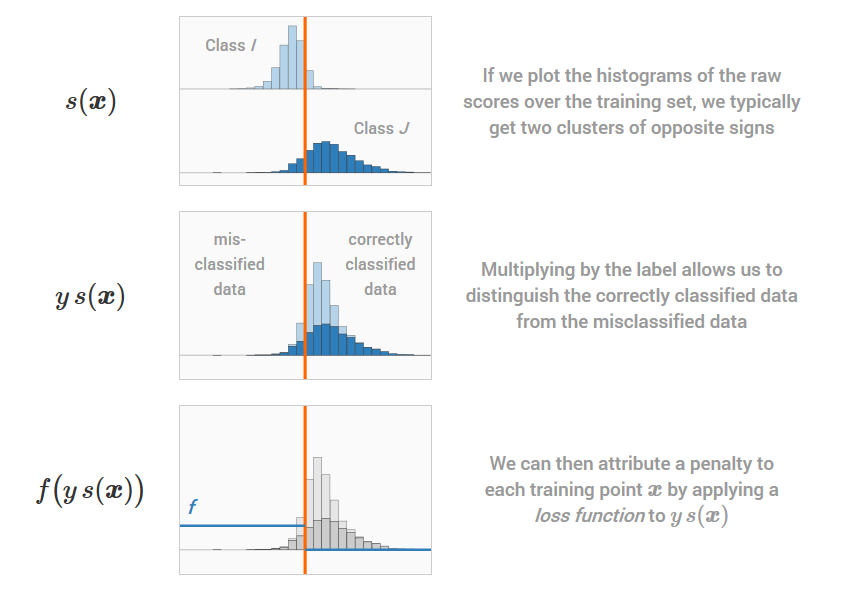}
\end{figure}

This leads to the notion of \emph{empirical risk} $R(\boldsymbol{w},b)$ for the classifier $\mathcal{C}$ defined as the average penalty over the training set $T$:
$$R(\boldsymbol{w},b) := \frac{1}{n}\sum_{(\boldsymbol{x},y) \in T}f\big(y\,s(\boldsymbol{x})\big)$$
In general, learning a linear classifier consists of finding a weight vector $\boldsymbol{w}$ and a bias $b$ minimising $R(\boldsymbol{w},b)$ for a well chosen loss function $f$.

In binary classification, three notable loss functions are:

\newpage

\begin{figure}[h!]
\includegraphics[width=\textwidth]{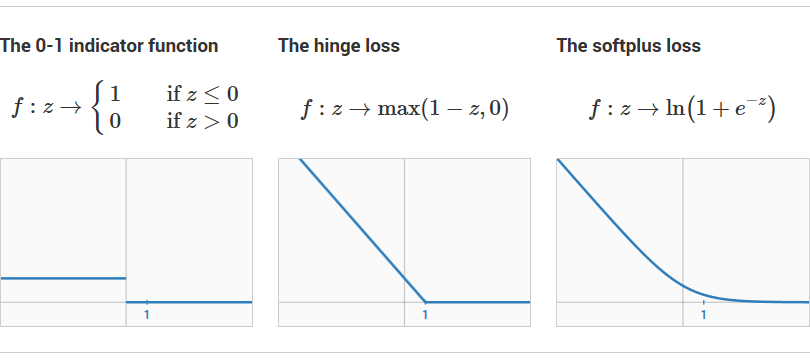}
\end{figure}

With the 0-1 indicator function, the empirical risk is simply the \emph{error rate} on $T$. In a sense, this is the optimal loss function as minimizing the error rate is often the desired objective in practice. Unfortunately, it is incompatible with gradient descent (there is no gradient to descend: the derivative is null everywhere).

This limitation is overcome in the hinge loss (used in SVM) and the softplus loss (used in logistic regression) by replacing the unit penalty on the misclassified data with a strictly decreasing penalty. Note that both the hinge loss and the softplus loss also penalize some correctly classified data in the neighbourhood of the boundary, effectively enforcing a safety margin.

{\textbf{The scaling parameter $\lVert\boldsymbol{w}\rVert$}}

An important point previously overlooked is that the signed distance $s(\boldsymbol{x})$ is scaled by the norm of the weight vector. If $d(\boldsymbol{x})$ is the actual \emph{signed Euclidean distance} between $\boldsymbol{x}$ and $\mathcal{C}$, we have:
$$d(\boldsymbol{x}):= \boldsymbol{\hat{w}} \!\cdot\! \boldsymbol{x} + b^\prime \quad\quad\quad \text{where} \quad\quad \textstyle \boldsymbol{\hat{w}} := \frac{\boldsymbol{w}}{\lVert\boldsymbol{w}\rVert}\,\quad b^\prime := \frac{b}{\lVert\boldsymbol{w}\rVert}$$
$$\text{and} \quad s(\boldsymbol{x}) = \lVert\boldsymbol{w}\rVert\,d(\boldsymbol{x})$$
Hence, the norm $\lVert\boldsymbol{w}\rVert$ can be interpreted as a scaling parameter for the loss function in the expression of the empirical risk:

\begin{figure}[h!]
  \centering
	\includegraphics[width=0.55\textwidth]{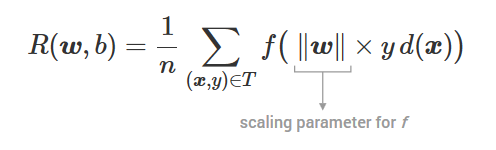}
\end{figure}

Let us define the \emph{scaled loss function} $f_{\lVert\boldsymbol{w}\rVert}:z \rightarrow f(\lVert\boldsymbol{w}\rVert \times z)$.

We observe that the 0-1 indicator function is invariant to rescaling while the hinge loss and the softplus loss are strongly affected.

\begin{figure}[h!]
\includegraphics[width=\textwidth]{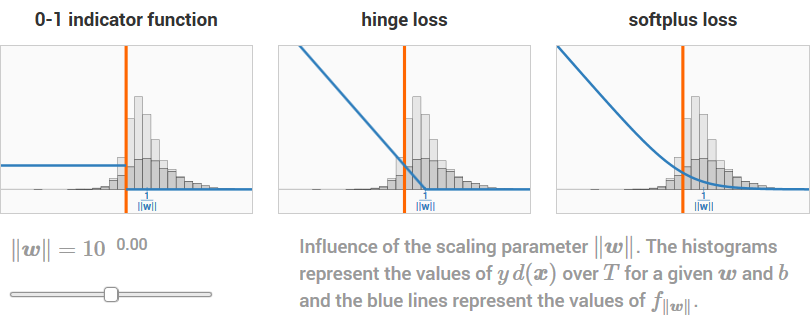}
\end{figure}

Remarkably, the hinge loss and the softplus loss behave in the same way for extreme values of the scaling parameter.

\vspace{0.5cm}

\begin{figure}[h!]
\includegraphics[width=0.85\textwidth]{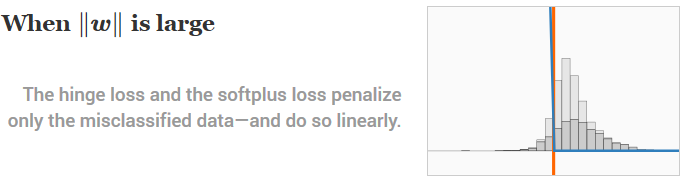}
\end{figure}

More precisely, both losses satisfy:\footnote{
  \underline{Hinge loss}
  \begin{align*}
    \max(1-\lVert\boldsymbol{w}\rVert\,y\,d(\boldsymbol{x}),0)& \quad\> \;=\; \quad\> \textstyle\lVert\boldsymbol{w}\rVert \, \max\left(\lVert\boldsymbol{w}\rVert^{-1}-y\,d(\boldsymbol{x}),0\right)\\
    &\;\underset{\lVert\boldsymbol{w}\rVert \to +\infty}{\approx}\; \lVert\boldsymbol{w}\rVert \, \max\left(-y\,d(\boldsymbol{x}),0\right)
  \end{align*}
	\phantom{aaa} \underline{Softplus loss}
  \begin{align*}
    \ln\left(1+e^{-\lVert\boldsymbol{w}\rVert\,y\,d(\boldsymbol{x})}\right) &\;\underset{\lVert\boldsymbol{w}\rVert \to +\infty}{\approx}\;
	  \begin{cases}
        -\lVert\boldsymbol{w}\rVert\,y\,d(\boldsymbol{x})  \quad & \text{if } y\,d(\boldsymbol{x}) \leq 0\\
        0                                                        & \text{if } y\,d(\boldsymbol{x}) > 0
      \end{cases}\\
	&\;\underset{\lVert\boldsymbol{w}\rVert \to +\infty}{\approx}\; \lVert\boldsymbol{w}\rVert \, \max\left(-y\,d(\boldsymbol{x}),0\right)
  \end{align*}}
$$f_{\lVert\boldsymbol{w}\rVert}\big(y\,d(\boldsymbol{x})\big) \,\underset{\lVert\boldsymbol{w}\rVert \to +\infty}{\approx}\, \lVert\boldsymbol{w}\rVert \, \max\left(-y\,d(\boldsymbol{x}),0\right)$$
For convenience, we name the set of misclassified data:
$$M := \{ (\boldsymbol{x},y) \in T \;|\; y\,d(\boldsymbol{x}) \leq 0 \}$$ 
and we can then write the empirical risk as:
$$R(\boldsymbol{w},b) \;\underset{\lVert\boldsymbol{w}\rVert \to +\infty}{\approx}\; \lVert\boldsymbol{w}\rVert \, \biggl(\frac{1}{n}\sum_{(\boldsymbol{x},y) \in M}\!\left(-\,y\,d(\boldsymbol{x})\right)\biggr)$$
This expression contains a term which we call the \emph{error distance}:
$$d_{\text{err}} := \frac{1}{n}\sum_{(\boldsymbol{x},y) \in M}\!\left(-\,y\,d(\boldsymbol{x})\right)$$
It is \emph{positive} and can be interpreted as the average distance by which each training sample is misclassified by $\mathcal{C}$ (with a null contribution for the correctly classified data). It is related---although not exactly equivalent---to the training error.\footnote{A small error distance $d_{\text{err}}$ does not \emph{guarantee} the training error $err_{\text{train}}$ to be small. In the worst case, when all the data lies on the boundary, $d_{\text{err}} = 0$ and $err_{\text{train}} = 100\%$.}

Finally we have:
$$\text{minimize: }R(\boldsymbol{w},b) \;\underset{\lVert\boldsymbol{w}\rVert \to +\infty}{\iff}\; \text{minimize: $d_{\text{err}}$}$$
In words, when $\lVert\boldsymbol{w}\rVert$ is large, \emph{minimizing the empirical risk for the hinge loss or the softplus loss is equivalent to minimizing the error distance, which is similar to minimizing the error rate on the training set}.

\vspace{0.5cm}

\begin{figure}[h!]
\includegraphics[width=0.85\textwidth]{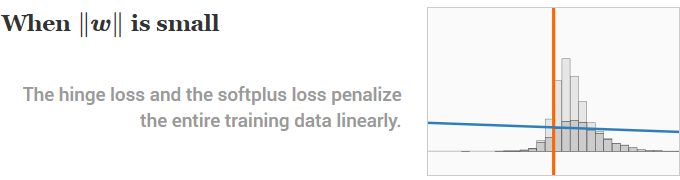}
\end{figure}

More precisely, both losses satisfy:\footnote{
  \underline{Hinge loss}
  $$\max(1-\lVert\boldsymbol{w}\rVert\,y\,d(\boldsymbol{x}),0) \;\underset{\lVert\boldsymbol{w}\rVert\ \to 0}{=}\; 1-\lVert\boldsymbol{w}\rVert\,y\,d(\boldsymbol{x})$$
  $$\alpha = 1 \quad\text{and}\quad \beta = 1$$
  \phantom{aaa} \underline{Softplus loss}
  $$\ln\left(1+e^{-\lVert\boldsymbol{w}\rVert\,y\,d(\boldsymbol{x})}\right) \;\underset{\lVert\boldsymbol{w}\rVert\ \to 0}{\approx}\; 
  \ln(2)-\frac{1}{2}\,\lVert\boldsymbol{w}\rVert\,y\,d(\boldsymbol{x})$$
  $$\alpha = \ln(2) \quad\text{and}\quad \beta = \frac{1}{2}$$}
$$f_{\lVert\boldsymbol{w}\rVert}\big(y\,d(\boldsymbol{x})\big) \,\underset{\lVert\boldsymbol{w}\rVert\ \to 0}{\approx}\, \alpha - \beta\;\lVert\boldsymbol{w}\rVert\,y\,d(\boldsymbol{x})$$
for some positive values $\alpha$ and $\beta$.

We can then write the empirical risk as:
$$R(\boldsymbol{w},b) \;\underset{\lVert\boldsymbol{w}\rVert\ \to 0}{\approx}\;\alpha - \beta\;\lVert\boldsymbol{w}\rVert \, \biggl(\frac{1}{n}\sum_{(\boldsymbol{x},y) \in T}y\,d(\boldsymbol{x})\biggr)$$
This expression contains a term which we call the \emph{adversarial distance}:
$$d_{\text{adv}} := \frac{1}{n}\sum_{(\boldsymbol{x},y) \in T}y\,d(\boldsymbol{x})$$
It is the mean distance between the images in $T$ and the classification boundary $\mathcal{C}$ (with a negative contribution for the misclassified images). It can be viewed as a measure of robustness to adversarial perturbations: when $d_{\text{adv}}$ is high, the number of misclassified images is limited and the correctly classified images are far from $\mathcal{C}$.

Finally we have:
$$\text{minimize: }R(\boldsymbol{w},b) \;\underset{\lVert\boldsymbol{w}\rVert \to 0}{\iff}\; \text{maximize: } d_{\text{adv}}$$
In words, when $\lVert\boldsymbol{w}\rVert$ is small, \emph{minimizing the empirical risk for the hinge loss or the softplus loss is equivalent to maximizing the adversarial distance, which can be interpreted as minimizing the phenomenon of adversarial examples.}

{\textbf{Closing remarks}}

In practice, the value of $\lVert\boldsymbol{w}\rVert$ can be controlled by adding a regularization term to the empirical risk, yielding the \emph{regularized loss}:
	
\begin{figure}[h!]
  \centering
	\includegraphics[width=0.55\textwidth]{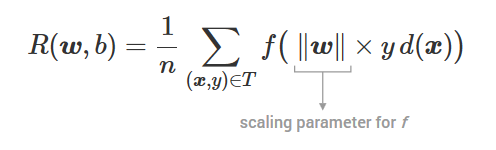}
\end{figure}

A small \emph{regularization parameter} $\lambda$ lets $\lVert\boldsymbol{w}\rVert$ grow unchecked while a larger $\lambda$ encourages $\lVert\boldsymbol{w}\rVert$ to shrink.

\begin{figure}[h!]
  \centering
	\includegraphics[width=\textwidth]{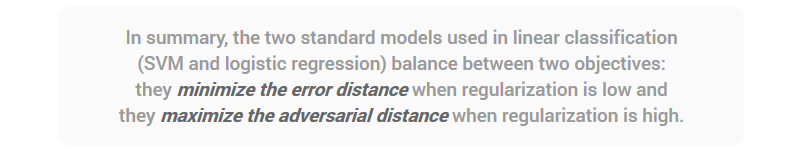}
\end{figure}

\vspace{0.5cm}

{\Large \textit{Adversarial Distance and Tilting Angle}}

The adversarial distance emerged in the previous section as a measure of robustness to adversarial perturbations. Rather conveniently, it can be expressed as a function of a single parameter: the angle between the classification boundary and the nearest centroid classifier.\newpage

If $T_I$ and $T_J$ are the restrictions of $T$ to the elements in $I$ and $J$ respectively, we can write: 
\begin{align*}
d_{\text{adv}} &= \frac{1}{n}\sum_{(\boldsymbol{x},y) \in T}y\,d(\boldsymbol{x})\\
				       &= \frac{1}{n}\,\biggl[\;\sum_{\boldsymbol{x} \in T_I}(-\boldsymbol{\hat{w}} \!\cdot\! \boldsymbol{x} - b^\prime) \;\;+\;\; \sum_{\boldsymbol{x} \in T_J}(\boldsymbol{\hat{w}} \!\cdot\! \boldsymbol{x} + b^\prime)\,\biggr]
\end{align*}
If $T_I$ and $T_J$ are balanced $\left(n = 2\,n_I = 2\,n_J\right)$:
\begin{align*}
d_{\text{adv}} &= -\frac{1}{2\,n_I}\sum_{\boldsymbol{x} \in T_I}\boldsymbol{\hat{w}} \!\cdot\! \boldsymbol{x} \;\;+\;\; \frac{1}{2\,n_J}\sum_{\boldsymbol{x} \in T_J}\boldsymbol{\hat{w}} \!\cdot\! \boldsymbol{x}\\
               &= \frac{1}{2}\;\boldsymbol{\hat{w}} \!\cdot\! \biggl[\biggl(\frac{1}{n_J} \sum_{\boldsymbol{x} \in T_J}\boldsymbol{x}\biggr) - \biggl(\frac{1}{n_I} \sum_{\boldsymbol{x} \in T_I}\boldsymbol{x}\biggr)\biggr]
\end{align*}
If $\boldsymbol{i}$ and $\boldsymbol{j}$ are the centroids of $T_I$ and $T_J$ respectively:
$$d_{\text{adv}} = \frac{1}{2}\;\boldsymbol{\hat{w}} \!\cdot\! (\boldsymbol{j}-\boldsymbol{i})$$
We now introduce the \emph{nearest centroid classifier}, which has unit normal vector $\boldsymbol{\hat{z}} = (\boldsymbol{j}-\boldsymbol{i})/\lVert\boldsymbol{j}-\boldsymbol{i}\rVert$:
$$d_{\text{adv}} =\frac{1}{2}\;\lVert\boldsymbol{j}-\boldsymbol{i}\rVert\;\boldsymbol{\hat{w}} \!\cdot\! \boldsymbol{\hat{z}}$$
Finally, we call the plane containing $\boldsymbol{\hat{w}}$ and $\boldsymbol{\hat{z}}$ the \emph{tilting plane} of $\mathcal{C}$ and we we call the angle $\theta$ between $\boldsymbol{\hat{w}}$ and $\boldsymbol{\hat{z}}$ the \emph{tilting angle} of $\mathcal{C}$:
$$d_{\text{adv}} = \frac{1}{2}\;\lVert\boldsymbol{j}-\boldsymbol{i}\rVert\;\cos(\theta)$$

This equation can be interpreted geometrically in the tilting plane:

\begin{figure}[h!]
\includegraphics[width=\textwidth]{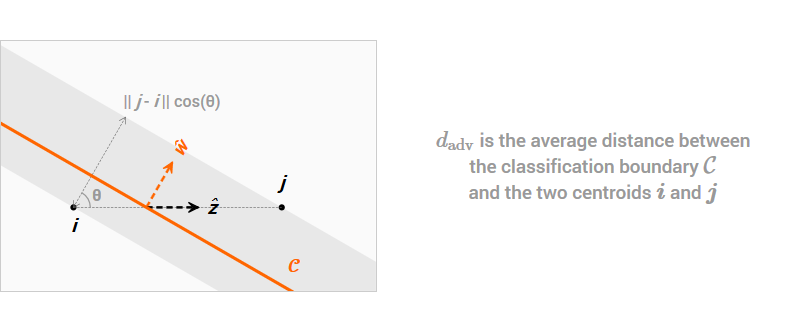}
\end{figure}

\newpage
On a given training set $T$ where the distance between the two centroids $\lVert\boldsymbol{j}-\boldsymbol{i}\rVert$ is fixed, $d_{\text{adv}}$ depends only on the tilting angle $\theta$. Two observations follow:
\begin{itemize}
\item The adversarial example phenomenon is minimized by the nearest centroid classifier ($\theta = 0$).\footnote{...and the classifiers parallel to it: $d_{\text{adv}}$ is independent of the bias $b$. This explains why the models defined by SVM are poorly adjusted when regularization is high, resulting in very high training and test errors (see for instance the classification of 1s vs 8s when $\lambda = 10^7$ in the next section).}
\item Adversarial examples can be arbitrarily strong when $\theta \to \pi/2$ \\
(as was the case with the classifier $\mathscr{L}_{\theta}$ in the toy problem section).
\end{itemize}

\vspace{0.5cm}

{\Large \textit{Example: SVM on MNIST}}

We now illustrate the previous considerations on binary classification of MNIST digits. For each possible pair of digit classes, we train multiple SVM models $(\boldsymbol{w}, b)$ for a regularization parameter $\lambda \in [10^{-1}, 10^7]$ using a training set of $3000$ images per class.\footnote{More precisely, we train for each pair of digit classes 81 models with a regularization parameter $\lambda = 10^{\alpha}$ with $\alpha$ ranging from $-1$ to $7$ by steps of $0.1$.}

We start by plotting the distributions of the distances $y\,d(\boldsymbol{x})$ between the training data and the boundary as a function of the regularization parameter $\lambda$ (grey histograms). We superimpose the loss function $f_{\lVert\boldsymbol{w}\rVert}$ as scaled after the convergence of each model (blue line).

\begin{figure}[h!]
\includegraphics[width=\textwidth]{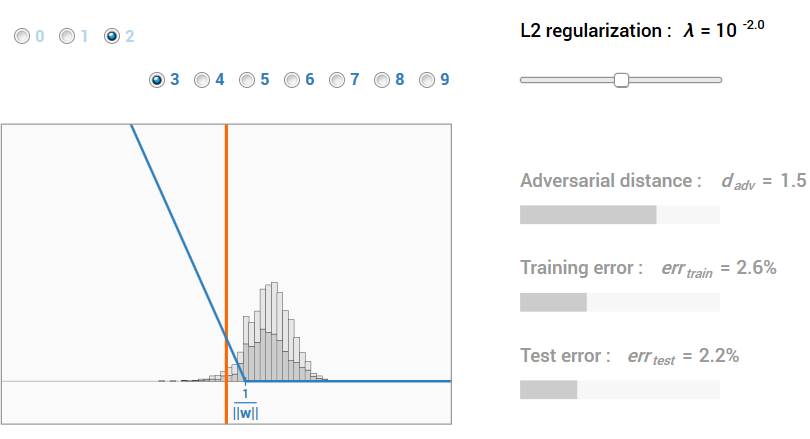}
\end{figure}

We see that the scaling of the hinge loss has a clear influence on the model obtained. Unfortunately, minimising the training error and maximizing the adversarial distance are conflicting goals: $err_{train}$ is minimized when $\lambda$ is small and $d_{adv}$ is maximized when $\lambda$ is large. Note that the test error is minimized for an intermediate level of regularization $\lambda_{\text{optimal}}$. When $\lambda < \lambda_{\text{optimal}}$, the classifier is \emph{overfitted} and when $\lambda > \lambda_{\text{optimal}}$, the classifier is \emph{underfitted}.

To get a better understanding of how the two objectives are balanced, we can look at the training data under a different point of view.
We first compute the unit weight vector $\boldsymbol{\hat{z}}$ of the nearest centroid classifier. Then for each SVM model $(\boldsymbol{w}, b)$, we compute the unit vector $\boldsymbol{\hat{n}}$ such that $(\boldsymbol{\hat{z}},\boldsymbol{\hat{n}})$ is an orthonormal basis of the tilting plane of $\boldsymbol{w}$.\footnote{We do this by using the Gram-Schmidt process: 
$$\boldsymbol{\hat{n}} = \frac{\boldsymbol{n}}{\lVert\boldsymbol{n}\rVert} \quad\quad \text{with} \quad\quad \boldsymbol{n} = \boldsymbol{w} - (\boldsymbol{w} \!\cdot\! \boldsymbol{\hat{z}})\,\boldsymbol{\hat{z}}$$} Finally, we project the training data in $(\boldsymbol{\hat{z}},\boldsymbol{\hat{n}})$:

\begin{figure}[h!]
\includegraphics[width=\textwidth]{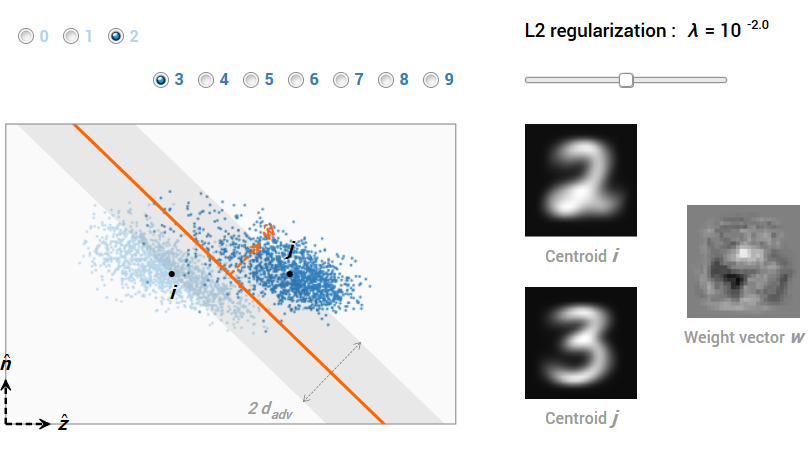}
\end{figure}

The horizontal direction passes through the two centroids and the vertical direction is chosen such that $\boldsymbol{w}$ belongs to the plane (the hyperplane boundary then appears as a line). Remark also that since $(\boldsymbol{\hat{z}},\boldsymbol{\hat{n}})$ is an orthonormal basis, the distances in this plane are actual pixel distances. To understand why the data points appear to be moving around when $\lambda$ varies, one needs to imagine the tilting plane rotating around $\boldsymbol{\hat{z}}$ in the 784-dimensional input space (thus showing a different section of the 784-dimensional training data for each value of $\lambda$).

For high regularization levels, the model is parallel to the nearest centroid classifier and the adversarial distance is maximized. As $\lambda$ decreases, the classification boundary improves its fit of the training data by tilting towards directions of low variance. Eventually, a small number of misclassified training samples is overfitted, resulting in a very small adversarial distance and a weight vector that is hard to interpret.

Finally, we can look at two representative images $\boldsymbol{x}$, $\boldsymbol{y}$ (one per class) and their mirror images $\boldsymbol{x}_m$, $\boldsymbol{y}_m$ for each model. Their projections in the tilting plane of $\boldsymbol{w}$ give a very intuitive picture of the adversarial example phenomenon in linear classification:

\begin{figure}[h!]
\includegraphics[width=\textwidth]{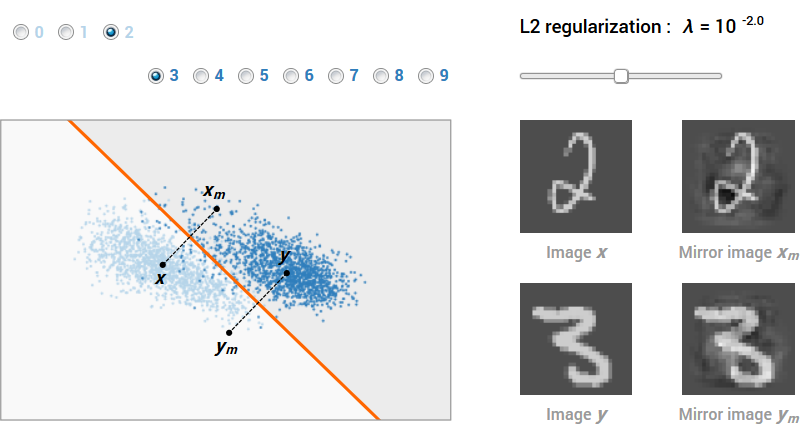}
\end{figure}

The model is sensitive to strong adversarial examples ($||\boldsymbol{x}_m-\boldsymbol{x}|| \to 0$ and $||\boldsymbol{y}_m-\boldsymbol{y}|| \to 0$) when the tilting angle approaches $\pi/2$. This is a symptom of strong overfitting, and whether it occurs or not depends on the difficulty to separate the two classes (compare for instance the classification of $7s$ versus $9s$ and the classification of $0s$ versus $1s$).

\noindent\makebox[\linewidth]{\textcolor{gray}{\rule{\paperwidth}{0.025cm}}}

\vspace{0.75cm}

{\Large Adversarial Examples in Neural Networks}

Thanks to the equivalence between adversarial distance and tilting angle, the linear case is simple enough to be visualized in the plane. In neural networks however, the class boundary is not flat and the adversarial distance cannot be reduced to a single parameter. Nonetheless, some similarities with the linear case remain.

\vspace{0.5cm}

{\Large \textit{First Step: a 2-Layer Binary Network}}

Let $\mathcal{N}$ be a 2-layer network with a single output defining a non-linear binary classifier in $\mathbb{R}^d$. The first layer of $\mathcal{N}$ is specified by a weight matrix $W_1$ and a bias vector $b_1$ and the second layer of $\mathcal{N}$ is specified by a weight vector $W_2$ and bias $b_2$. We assume that the two layers are separated by a layer $\phi$ of rectified linear units applying the function $z \to \max(0,z)$ element-wise. For an image $x$ in $\mathbb{R}^d$, we call the \emph{raw score} of $x$ through $\mathcal{N}$ the value:
$$s(x) := W_2\;\phi(W_1\,x + b_1) + b_2$$ 
Similarly to the linear case, the \emph{empirical risk} on $T$ for a \emph{loss function} $f$ can be written:
$$R(W_1,b_1;W_2,b_2) := \frac{1}{n}\sum_{(x,y) \in T}f\big(y\,s(x)\big)$$
and training $\mathcal{N}$ consists in finding $W_1$, $b_1$, $W_2$ and $b_2$ minimizing $R$ for a well chosen $f$.

$\phi$ is piecewise linear and around each image $x$ there is a local linear region $\mathcal{L}_x$ within which:
$$\phi(W_1 \, x + b_1) = W_1^x \, x + b_1^x$$
where $W_1^x$ and $b_1^x$ are obtained by zeroing some lines in $W_1$ and $b_1$ respectively.\footnote{More precisely, the $i^{th}$ lines in $W_1^x$ and $b_1^x$ are:
$$\left(W_1^x\,,\,b_1^x\right)_i = 
\begin{cases}
  \left(W_1\,,\,b_1\right)_i  \quad & \text{if } \left(W_1\,x + b_1\right)_i > 0\\
  0                                 & \text{if } \left(W_1\,x + b_1\right)_i \leq 0
\end{cases}$$}
Within $\mathcal{L}_x$, the raw score can thus be written:
$$s(x) = W_2 W_1^x \, x + W_2 b_1^x + b_2$$
This can be seen as the raw score of a local linear classifier $\mathcal{C}_x$ and our analysis of the linear case then applies almost without modifications. First, we observe that $s(x)$ is a scaled distance. If $d(x)$ is the actual \emph{signed Euclidean distance} between $x$ and $\mathcal{C}_x$, we have:
$$s(x) = \lVert W_2 W_1^x \rVert\,d(x)$$
The norm $\lVert W_2 W_1^x \rVert$ can then be interpreted as a scaling parameter for the loss function (the scaling is now local, dependent on $x$). One simple way to control all the local scalings simultaneously is by adding an L2 regularization term to the empirical risk acting on the norms $\lVert W_1 \rVert$ and $\lVert W_2 \rVert$ independently (remember that the weights in $W_1^x$ are a subset of the weights in $W_1$). With gradient descent, this is equivalent to decaying the weights $W_1$ and $W_2$ at every iteration. More precisely, for a learning rate $\eta$ and a decaying factor $\lambda$, the \emph{weight decay update} is:
$$W_1 \leftarrow W_1 - \eta\,\lambda\,W_1 \quad \text{and} \quad W_2 \leftarrow W_2 - \eta\,\lambda\,W_2$$
\begin{itemize}
\item \textbf{With a small decaying factor$\;\lambda$,} the scaling parameter $\lVert W_2 W_1^x \rVert$ is allowed to grow unrestricted and the loss penalizes only the misclassified data. Minimizing the empirical risk is then equivalent to minimizing the error on the training set.
\item \textbf{As the decaying factor$\;\lambda\;$increases,} the scaling parameter $\lVert W_2 W_1^x \rVert$ decreases and the loss starts penalizing more and more of the correctly classified data, pushing it further away from the boundary. Under this light, L2 weight decay can be seen as a form of \emph{adversarial training}.
\end{itemize}

\begin{figure}[t!]
\includegraphics[width=\textwidth]{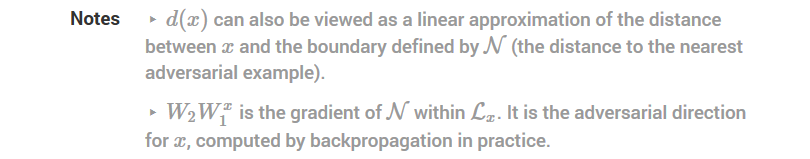}
\end{figure}

\begin{figure}[h!]
\includegraphics[width=\textwidth]{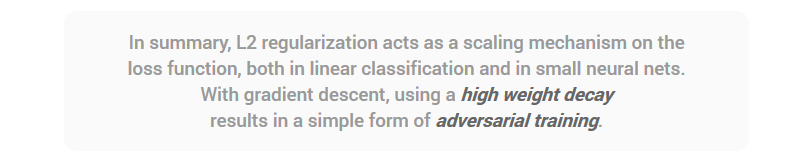}
\end{figure}

\vspace{0.5cm}

{\Large \textit{Second Step: General Case}}

The previous analysis can be generalized to more layers and even to non-piecewise-linear activation functions. The important observation is that we always have:
$$s(x) = \lVert \nabla_{\!x} \, s \rVert\, d(x)$$
Where $\nabla_{\!x} \, s$ is the gradient of the raw score on $x$, and $d(x)$ is a linear approximation of the distance between $x$ and the boundary defined by the network. The norm $\lVert \nabla_{\!x} \, s \rVert$ then constitutes a scaling parameter for the loss function which can be controlled with weight decay.
	
This idea can also be extended beyond binary classification. In the multiclass case, the raw score becomes a vector whose elements are called the \emph{logits}. Each logit $s_i(x)$ is then transformed into a \emph{probability} $p_i(x)$ by applying the softmax function:
$$p_i(x) := \text{softmax}_i(s(x)) := \frac{e^{s_i(x)}}{\displaystyle \sum_j \textstyle e^{s_j(x)}}$$
For an image/label pair $(x,y)$, the probability associated with the correct class is $p_y(x)$. The \emph{log-likelihood} loss function encourages it to come close to $1$ by attributing the following penalty to $(x,y)$:

\begin{figure}[h!]
\includegraphics[width=\textwidth]{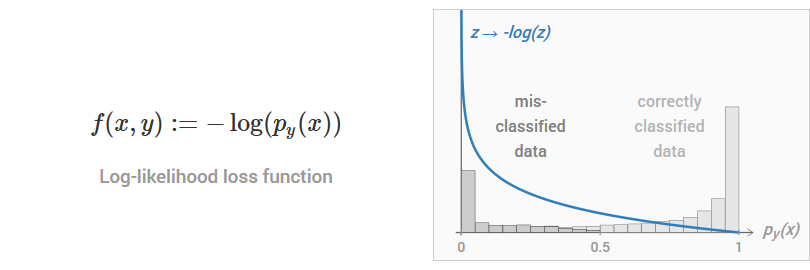}
\end{figure}

Now, varying weight decay influences the scaling of the logits, effectively acting as a \emph{temperature} parameter for the softmax function.\footnote{We can redefine the softmax function as:
$$\text{softmax}_i(s(x),t) := \frac{e^{\frac{s_i(x)}{t}}}{\displaystyle \sum_j \textstyle e^{\frac{s_j(x)}{t}}}$$
where $t$ is the temperature parameter. It was introduced in the context of network distillation \citep{hinton2015distilling} and controls the softness of the softmax function.} When weight decay is very low, the probability distributions generated are close to one-hot encodings ($p_y(x) \approx 0 \text{ or } 1$) and only the misclassified data produces non-zero penalties. With higher weight decay however, the probability distributions generated become smoother and the correctly classified data participates to the training, thus preventing overfitting.

In practice, a number of observations suggest that modern deep networks are under-regularized:
\begin{enumerate}[topsep=0pt]
\item They are often poorly calibrated and produce overconfident predictions \citep{guo2017calibration}.
\item They often converge to zero training error, even on a random labelling of the data \citep{zhang2016understanding}.
\item They are often vulnerable to linear attacks of small magnitude \citep{goodfellow2014explaining}.
\end{enumerate}

\vspace{0.5cm}

{\Large \textit{Example: LeNet on MNIST}}

Is it possible to regularize a neural network against adversarial examples by only using weight decay? The idea is simple enough and has been considered before: Goodfellow et al. \citep{goodfellow2014explaining} have observed that adversarial training is \emph{"somewhat similar to L1 regularization"} in the linear case. However, the authors reported that when training maxout networks on MNIST, an L1 weight decay coefficient of $0.0025$ \emph{"was too large, and caused the model to get stuck with over $5\%$ error on the training set. Smaller weight decay coefficients permitted successful training but conferred no regularization benefit."} We put the idea to the test once again and our observations are more nuanced. If using a high weight decay is clearly not the panacea, we found that it does help reduce the adversarial examples phenomenon, at least in simple setups.

Consider LeNet on MNIST (10-class problem). We use the baseline MatConvNet \citep{vedaldi2015matconvnet} implementation with the following architecture:

\begin{figure}[h!]
  \centering
	\makebox[\textwidth][c]{\includegraphics[width=1.3\textwidth]{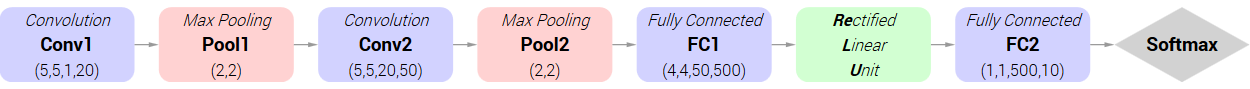}}
\end{figure}

We train one version of this network with a \emph{low} weight decay of $10^{-4}$ and one version with a \emph{high} weight decay of $10^{-1}$ (we refer to the two versions as $\text{LeNet}_{\text{\emph{low}}}$ and $\text{LeNet}_{\text{\emph{high}}}$ respectively). We keep all the other parameters fixed: we train for $50$ epochs, use a batch size of $300$, a learning rate of $0.0005$ and a momentum of $0.9$.

We can make several observations. To start, let's plot the training and test errors for the two networks as a function of the epoch.

\begin{figure}[h!]
  \centering
	\makebox[\textwidth][c]{\includegraphics[width=1.3\textwidth]{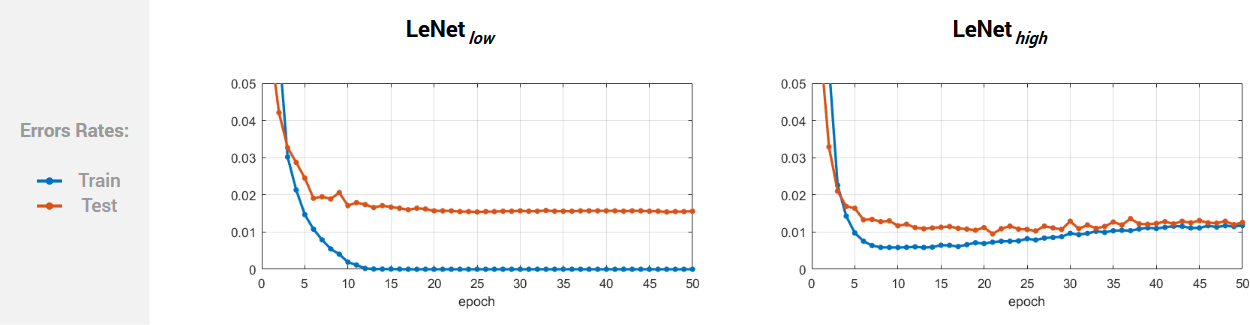}}
\end{figure}

We see that $\text{LeNet}_{\text{\emph{high}}}$ is less overfitted (the train and test errors are approximately equal at the end of the training) and performs slightly better than $\text{LeNet}_{\text{\emph{low}}}$ (final test error of $1.2\%$ versus $1.6\%$).

We can also inspect the weights that have been learnt. Below, we compute their root mean square value (RMS) and show a random selection of filters for each convolutional layer.

\begin{figure}[h!]
  \centering
	\makebox[\textwidth][c]{\includegraphics[width=1.3\textwidth]{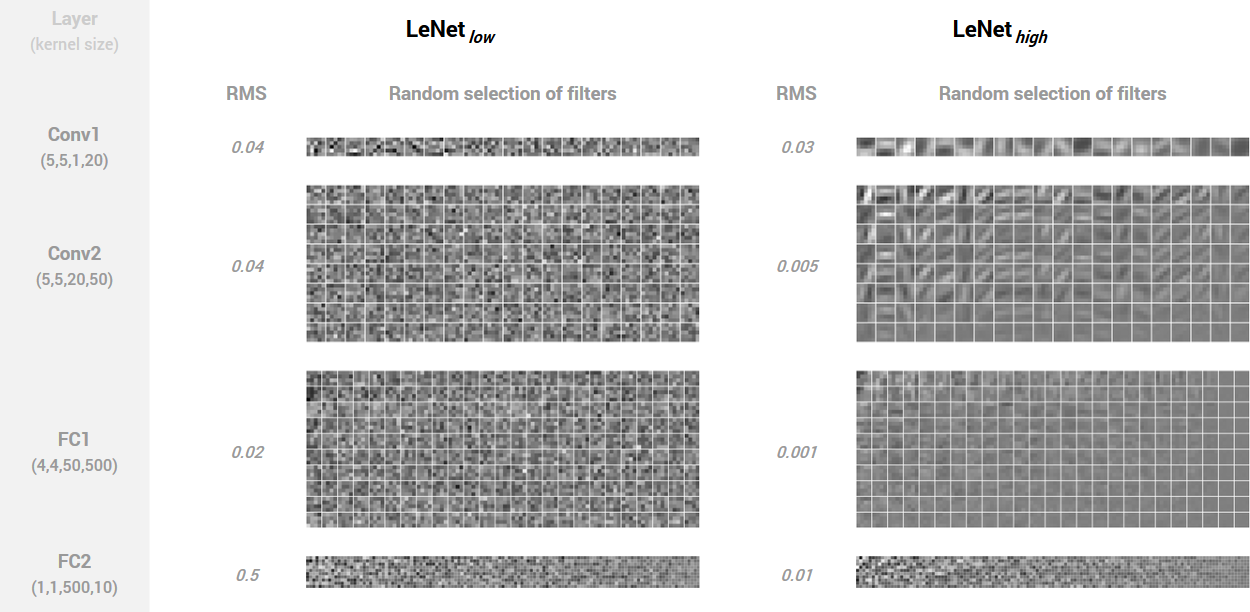}}
\end{figure}

As expected, the weights learnt with a higher weight decay have a much lower RMS. The filters of $\text{LeNet}_{\text{\emph{high}}}$ are also smoother than the filters of $\text{LeNet}_{\text{\emph{low}}}$ (see the presence of clean edge detectors in $\text{Conv1}$ and $\text{Conv2}$) and their magnitudes vary more within each convolutional layer (see the presence of uniformly gray filters in $\text{Conv2}$ and $\text{FC1}$).

Finally, let's submit the two networks to the same visual evaluation: for a random instance of each digit, we generate a high confidence adversarial example targeted to perform a cyclic permutation of the labels $0 \to 1$, $1 \to 2$, ..., $9 \to 0$. Specifically, each adversarial example is generated by performing gradient ascent on the probability of the desired label until the median value of $0.95$ is reached.\footnote{On each network, we set the temperature $t$ of the softmax layer such that the median classification score over the test set is 0.95.
For a target label $l$, the gradient $\nabla_{\!x}\, p_l(x)$ (computed by backpropagation) gives the direction of steepest ascent towards $l$. Here, we generated each adversarial example $x^\prime$ by iterating the following update rule:
$$x^\prime = \text{clip}_{[0,255]} \left(x^\prime + 1.0 \times \frac{\nabla_{\!x^\prime}\, p_l(x^\prime)}{\lVert \nabla_{\!x^\prime}\, p_l(x^\prime) \rVert}\right)$$
until $p_l(x^\prime) = 0.95$.} We show below the $10$ original images $\text{OI}$ with their corresponding adversarial examples $\text{AE}$ and adversarial perturbations $\text{Pert}$ for the two networks.

\begin{figure}[h!]
  \centering
	\makebox[\textwidth][c]{\includegraphics[width=1.3\textwidth]{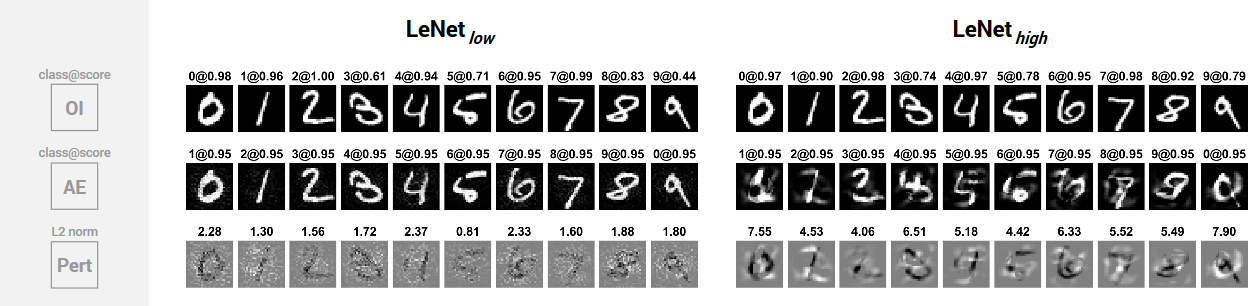}}
\end{figure}

We see that $\text{LeNet}_{\text{\emph{high}}}$ is less susceptible to adversarial examples than $\text{LeNet}_{\text{\emph{low}}}$: the adversarial perturbations have higher L2 norms and are more meaningful for human observers.

\noindent\makebox[\linewidth]{\textcolor{gray}{\rule{\paperwidth}{0.025cm}}}

\vspace{0.75cm}

{\Large Thoughts Moving Forward}

Despite the widespread interest it has generated for several years now, and despite its significance for the field of machine learning both in theory and in practice, the adversarial example phenomenon has so far retained much of its intrigue. Our main goal here was to provide a clear and intuitive picture of the phenomenon in the linear case, hopefully constituting a solid base from which to move forward. Incidentally, we showed that L2 weight decay plays a more significant role than previously suspected in a small neural net on MNIST.

Unfortunately, the story gets more complicated with deeper models on more sophisticated datasets. In our experience, the more non-linear the model becomes and the less weight decay seems to be able to help. This limitation may be superficial and it is perhaps worth exploring the ideas introduced here a bit further (for example, we should probably pay more attention to the scaling of the logits during training). Or the high non-linearity of deep networks might constitute a fundamental obstacle to the type of first-order adversarial training that L2 regularization implements. Our feeling is that a truly satisfying solution to the problem will likely require profoundly new ideas in deep learning.

\newpage
\bibliographystyle{unsrt}
\bibliography{biblio}

\end{document}